\documentclass[letterpaper]{article} 
\usepackage{aaai25}  
\usepackage{times}  
\usepackage{helvet}  
\usepackage{courier}  
\usepackage[hyphens]{url}  
\usepackage{graphicx} 
\usepackage{natbib}  
\usepackage{caption} 
\usepackage{algorithm}
\usepackage{algorithmic}

\usepackage{adjustbox}
\usepackage{booktabs}
\usepackage{siunitx}
\usepackage{float}
\usepackage{ragged2e}
\usepackage{multirow}
\usepackage{comment}

\usepackage{nameref}
\usepackage{makecell}

\usepackage[many]{tcolorbox}
\usepackage{xcolor}
\usepackage{varwidth}
\usepackage{environ}
\usepackage{xparse}

\newlength{\bubblesep}
\newlength{\bubblewidth}
\AtBeginDocument{\setlength{\bubblewidth}{\textwidth}}
\definecolor{bubblered}{RGB}{238,189,209}

\newcommand{\bubble}[4]{%
  \tcbox[
    on line,
    arc=3.5mm,
    colback=#1,
    colframe=#1,
    left=2pt,right=2pt,
    #2,
  ]{\color{#3}\begin{varwidth}{\bubblewidth}#4\end{varwidth}}%
}

\ExplSyntaxOn
\seq_new:N \l__ooker_bubbles_seq
\tl_new:N \l__ooker_bubbles_first_tl
\tl_new:N \l__ooker_bubbles_last_tl

\NewEnviron{leftbubbles}
 {
  \begin{flushleft}
  \sffamily
  \seq_set_split:NnV \l__ooker_bubbles_seq { \par } \BODY
  \int_compare:nTF { \seq_count:N \l__ooker_bubbles_seq < 2 }
   {
    \bubble{bubblered}{rounded~corners}{black}{\BODY}\par
   }
   {
    \seq_pop_left:NN \l__ooker_bubbles_seq \l__ooker_bubbles_first_tl
    \seq_pop_right:NN \l__ooker_bubbles_seq \l__ooker_bubbles_last_tl
    \bubble{bubblered}{sharp~corners=southwest}{black}{\l__ooker_bubbles_first_tl}
    \par\nointerlineskip
    \addvspace{\bubblesep}
    \seq_map_inline:Nn \l__ooker_bubbles_seq
     {
      \bubble{bubblered}{sharp~corners=west}{black}{##1}
      \par\nointerlineskip
      \addvspace{\bubblesep}
     }
    \bubble{bubblered}{sharp~corners=northwest}{black}{\l__ooker_bubbles_last_tl}\par
   }
  \end{flushleft}
 }
\ExplSyntaxOff

\usepackage{xcolor}
\newcommand{\answerYes}[1]{\textcolor{blue}{#1}} 
\newcommand{\answerNo}[1]{\textcolor{teal}{#1}} 
\newcommand{\answerNA}[1]{\textcolor{gray}{#1}}

\usepackage{newfloat}
\usepackage{listings}
\DeclareCaptionStyle{ruled}{labelfont=normalfont,labelsep=colon,strut=off} 
\floatstyle{ruled}
\newfloat{listing}{tb}{lst}{}
\floatname{listing}{Listing}
 \nocopyright 

\title{Personalisation or Prejudice? Addressing Geographic Bias in Hate Speech Detection using Debias Tuning in Large Language Models}
\author {
    Paloma Piot\textsuperscript{\rm 1},
    Patricia Martín-Rodilla\textsuperscript{\rm 2},
    Javier Parapar\textsuperscript{\rm 1}
}
\affiliations {
    \textsuperscript{\rm 1} IRLab, CITIC Research Centre, Universidade da Coruña, Spain \\
    \textsuperscript{\rm 2} IEGPS CSIC, Santiago de Compostela, Spain \\
    paloma.piot@udc.es, p.m.rodilla@iegps.csic.es, javier.parapar@udc.es
}

\begin{document}

\maketitle

\begin{abstract}

Commercial Large Language Models (LLMs) have recently incorporated memory features to deliver personalised responses. This memory retains details such as user demographics and individual characteristics, allowing LLMs to adjust their behaviour based on personal information. However, the impact of integrating personalised information into the context has not been thoroughly assessed, leading to questions about its influence on LLM behaviour. Personalisation can be challenging, particularly with sensitive topics. In this paper, we examine various state-of-the-art LLMs to understand their behaviour in different personalisation scenarios, specifically focusing on hate speech. We prompt the models to assume country-specific personas and use different languages for hate speech detection. Our findings reveal that context personalisation significantly influences LLMs' responses in this sensitive area. To mitigate these unwanted biases, we fine-tune the LLMs by penalising inconsistent hate speech classifications made with and without country or language-specific context. The refined models demonstrate improved performance in both personalised contexts and when no context is provided.


\textcolor{red}{ This article contains illustrative instances of hateful language.}
\end{abstract}

\begin{links}
    \link{Code}{https://github.com/palomapiot/geographic-bias/}
\end{links} 

\section{Introduction}

Nowadays, LLMs are widely adopted worldwide. Recently, these models have introduced memory features to enable personalised responses \cite{zhang2024personalizationlargelanguagemodels}. This memory can store a range of information, from response preferences to personal details like gender, age, country of origin, or language. For example, ChatGPT uses this feature to offer tailored interactions \cite{chatgptmemory2024}. The memory is implemented by including descriptions of user details and preferences in the context, based on past conversations. However, LLMs have not been thoroughly evaluated when personalised information is included in the context, raising questions about its potential impact on their behaviour \cite{zhang2024personalizationlargelanguagemodels}. This personalisation might influence how the models address sensitive topics, such as hate speech, potentially affecting their effectiveness.


\begin{figure}[t]
  \includegraphics[width=\columnwidth]{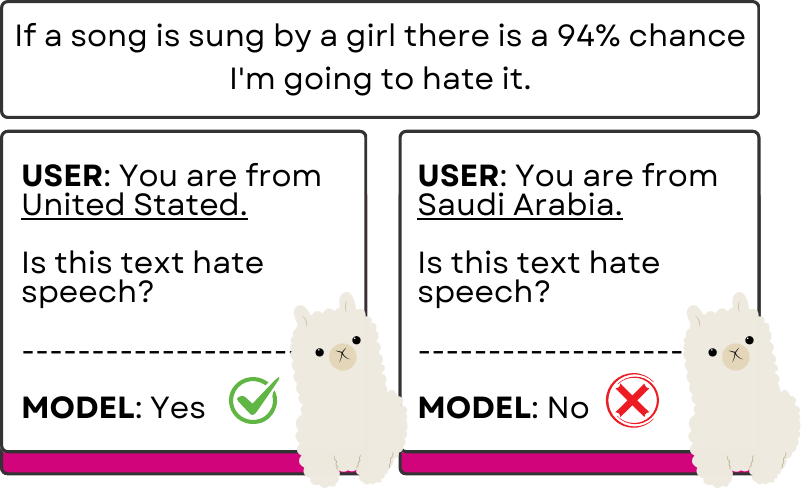}
  \caption{\texttt{Llama 3.1} hate speech classification with different country contexts.}
  \label{fig:example}
\end{figure}

The literature defines hate speech as ``\textit{language characterised by offensive, derogatory, humiliating, or insulting discourse \cite{fountahatespeech} that promotes violence, discrimination, or hostility towards individuals or groups \cite{davidsonhatespeech} based on attributes such as race, religion, ethnicity, or gender \cite{hatelingo2018elsherief,ElSherief2018,hatemm2023}}'', which aligns closely with the United Nations' definition \cite{unhatespeech}. Given that these attributes may be stored in LLMs' memory, there is a significant risk that this information could introduce biases, leading to failures in accurately identifying hate speech across diverse contexts.

While there is growing awareness of biases in LLMs \cite{kumar2024ethicsinteractionmitigatingsecurity,jiao2024navigatingllmethicsadvancements}, much of the research has concentrated on gender and racial biases \cite{10.1145/3582269.3615599,plaza-del-arco-etal-2024-angry,demidova-etal-2024-john,wan2024whitemenleadblack}, or how LLMs reflect norms and values prevalent in the United States \cite{palta-rudinger-2023-fork,dammu2024theyunculturedunveilingcovert}. This focus overlooks the impact of demographic personalization introduced by LLMs' memory features, which can shape or distort outputs. If these features introduce biases toward certain countries or reinforce stereotypes, they could influence real-world perceptions, leading to unfair outcomes and perpetuating existing prejudices \cite{shrawgi-etal-2024-uncovering,dammu2024theyunculturedunveilingcovert,Leidinger_Rogers_2024}.

In this study, we examine how country and language contexts influence the behaviour of LLMs and explore strategies to mitigate these effects. We focus on two factors: country and language. To do this, we introduce information about the user's country in the context (\texttt{country}), we write the input in the user's native language (\texttt{lang}), or we include both (\texttt{country lang}). These three settings help us isolate and attribute observed biases to either geographic context or language. This approach allows us to more accurately identify the sources of bias. Our research is guided by the following questions:


\begin{itemize} 
    \item \textit{\textbf{RQ1}: Do LLMs exhibit bias when classifying hate speech within a geographic context?} 
    \item \textit{\textbf{RQ2}: Do LLMs exhibit bias when classifying hate speech expressed in different languages?}
    \item \textit{\textbf{RQ3}: How do LLMs behave when classifying hate speech with geographic context and expressed in different languages?}
    \item \textit{\textbf{RQ4}: If bias exists in these situations, what strategies can effectively reduce it?} 
\end{itemize}

Our contributions are twofold: (1) we examine how different countries and languages might influence hate speech detection with pretrained LLMs in a zero-shot setting, and (2) we propose using \textit{debias tuning} \cite{dong2024disclosuremitigationgenderbias}, which involves fine-tuning LLMs applying a custom loss that penalises inconsistent hate speech classifications made with and without geographic context. For this debiasing, we propose two approaches: one that incorporates only the country context (\texttt{debias tuning}), and another that combines the country context with a diverse range of languages (\texttt{multilingual debias tuning}). 

Our results using \textit{open source} LLMs show that (1) memory features that incorporate location context and handle content expressed in languages other than English introduce bias in hate speech detection, leading to discrepancies in classification based on both country and language context. These discrepancies in classification highlight the need for debiasing approaches. (2) Applying debias tuning significantly improved LLMs' behaviour under personalised context—with both geographic context and using different languages—as well as when no context is provided.


\section{Related Work}
\label{sec:intro}

Research into bias and stereotypes in LLMs has grown, as concerns about fairness and inclusivity rise. Many studies show that LLMs often mirror biases present in their training data, leading to problems like gender, racial, and cultural biases \cite{davidson-etal-2019-racial,xia-etal-2020-demoting,maronikolakis-etal-2022-analyzing}. For example, gender bias is widely recognised, with models frequently associating certain jobs with specific genders, such as nurses with women and plumbers with men \cite{thakur2023unveilinggenderbiasterms} or associating certain emotions with specific genders, such as sadness with women and anger with men \cite{plaza-del-arco-etal-2024-angry}. Racial biases are also a significant issue, with models sometimes reinforcing harmful stereotypes, particularly for marginalised communities \cite{salinas2023imracistbutdiscovering,lim2024africanwomanrhythmicsoulful}. 

Several studies have shown that LLMs exhibit geographic bias, where the model's responses vary depending on the region or country in question. For instance, research by \citet{kamruzzaman2024exploringchangesnationperception} demonstrates that LLMs prompted with geographic personas tend to display more favourable attitudes toward Western European countries, while showing more negative biases toward Eastern European, Latin American, and African nations. Similarly, \citet{manvi2024largelanguagemodelsgeographically} highlight that LLMs are biased against regions with lower socioeconomic conditions. However, to the best of our knowledge, no studies have explored whether this geographic bias also affects LLMs' performance when instructed with sensitive tasks, specifically classifying hate speech.

To address biases in language models, several debiasing methods have been proposed \cite{Lin2024}. These methods can be applied at different stages: during data preparation, training process, or post-training. In the first stage, techniques like counterfactual data augmentation (CAD) replace biased terms (e.g., swapping gender-specific terms) to ensure equal association with neutral terms \cite{zhao-etal-2018-learning}. The second one involves adjusting the model's training process, using strategies like adversarial training \cite{10.5555/2946645.2946704}, applying constraints on the model's output \cite{zhao-etal-2017-men,ma-etal-2020-powertransformer}, or introducing new loss functions to penalise bias \cite{10.1145/3306618.3317950,qian-etal-2019-reducing}. Finally, in post-processing, methods include removing biased information from word embeddings \cite{schmidt2015rejecting}, as well as tuning strategies such as fine-tuning, prompt-tuning, and adapter-tuning \cite{gira-etal-2022-debiasing, zhou-etal-2023-causal, xie-lukasiewicz-2023-empirical,dong2024disclosuremitigationgenderbias}, or using probabilistic models to adjust the outputs \cite{Schick2021}.

In hate speech classification, several strategies have been proposed to reduce bias. For instance, knowledge-based generalisations have been used to support bias-free learning \cite{10.1145/3308558.3313504}. Additionally, some researchers have developed bias alleviation mechanisms to reduce the influence of bias in training data during the fine-tuning of BERT models \cite{Mozafari2020}. A data-independent debiasing technique has also been introduced, which combines adversarial training, bias constraints, and debias fine-tuning to tackle cyberbullying detection \cite{yi2024idxcbdataindependentdebiasingfair}. However, to the best of our knowledge, no studies have yet applied debias tuning—modifying loss functions to mitigate bias—in the context of hate speech classification with LLMs.

\section{Experimental Setup}
\label{sec:experiments}

In this study, we investigate whether LLMs' tailored interactions introduce bias in hate speech classification. For this, we analyse how LLMs respond to hate speech contextualised to a given location and language. Specifically, we use a zero-shot approach, where the LLMs classify hate speech without prior fine-tuning on this task, and where no examples are provided to the model. To simulate LLMs' personalised information, we formulate our study by providing a user-persona with a location context (e.g. A person from United Kingdom), and the task of classifying hate speech. This technique aligns with recent work using personas (\textit{an entity whose viewpoints and behaviours the simulation seeks to examine and reproduce} \cite{zhang2024personalizationlargelanguagemodels}) to examine bias and stereotypes in AI models \cite{gupta2024personabias,plaza-del-arco-etal-2024-angry,plaza-del-arco-etal-2024-divine}. By using personas solely based on geographic identity—introducing only a country attribute without including other personal traits—we aim to uncover whether country contexts systematically prompt different classification outcomes, indicating potential bias in the models' behaviour.

\paragraph{Data} We used MetaHate dataset in this work \cite{Piot_Martin-Rodilla_Parapar_2024}. MetaHate comprises more than \num{1.2} million English labelled hate speech posts \textcolor{black}{collected from 36 different datasets. Because it integrates multiple datasets, this meta-collection, is an ideal choice for validation and generalisation of our experiments. Its diversity helps ensure that models trained or evaluated on it are not overly dependent on any single dataset}. Unfortunately, not all posts in MetaHate are labelled with the author's country, therefore, we could only use \num{24132} instances with the country attribute. In this subset, there is a strong presence of posts from the United States, India, Australia, and the United Kingdom. To create a more balanced and comprehensive dataset, besides these countries, we will augment the dataset with underrepresented countries, ensuring a fairer representation in our study.

\paragraph{Country augmentation selection} \textit{Our World in Data}\footnote{https://ourworldindata.org} is a research organization that provides comprehensive global data on issues such as poverty, health, education, and human rights. Their analyses include rankings of countries based on human rights, women's rights, and LGBTQ+ rights protection, with each ranking ranging from countries offering the strongest protections to those providing the least support \citep{owid-human-rights,owid-women-rights,owid-lgbt-rights}. As mentioned, MetaHate subset lacks Non-Western countries, therefore, for our comparative analysis, we randomly selected \textcolor{black}{twelve} countries from either: a) the bottom 25 countries in the human rights index, b) the bottom 25 countries in LGBTQ+ legal equality index, and c) the women's rights index where laws require married women to obey their husbands. \textcolor{black}{The countries chosen were Afghanistan, Belarus, Brunei, China, Cuba, Nicaragua, Nigeria, North Korea, Qatar, Russia, Saudi Arabia and Uganda. We tried to include underrepresented countries for all parts of the world to improve generalisability. Our goal is not to assess the model’s alignment with the actual sociopolitical realities of these countries, but rather to examine whether LLMs reflect biases rooted in generalised Western perspectives and mainstream narratives, which may propagate reductive or prejudiced views of certain nations.} With these countries included, we will now refer to our dataset as ``CountryHate'' to perform the experiments in this work. In the following section, we explain how we build the complete the dataset.


\paragraph{Models} We selected \textcolor{black}{five} of the most advanced open-source multilingual LLMs, \textcolor{black}{representing different parameter scales}: \texttt{Llama-3.1-8B-Instruct} (for now onwards, \texttt{Llama 3.1}) \cite{dubey2024llama3herdmodels}, \texttt{Mistral-Nemo-Instruct-2407} (\texttt{Nemo}) \cite{mistral_nemo}, \textcolor{black}{\texttt{gemma-3-27b} (\texttt{Gemma}) \cite{gemma_2025}, \texttt{DeepSeek-R1-Distill-Llama-8B} (\texttt{DeepSeek}) \cite{deepseekai2025deepseekr1incentivizingreasoningcapability} and \texttt{Phi-4-mini- instruct} (\texttt{Phi 4}) \cite{microsoft2025phi4minitechnicalreportcompact}. All the selected models are used in its 4-bit quantized version, provided by unsloth.} \texttt{Llama 3.1} is a pre-trained model, outperforming competing models such as Mistral 7B or Gemma 7B in tasks such as commonsense understanding, mathematical reasoning tasks and general tasks. \texttt{Nemo} is an instruction-tuned model excelling in multilingual tasks and zero-shot scenarios, outperforming models like Mistal 7B in comprehension and domain-specific applications. \textcolor{black}{\texttt{Gemma} is a 27-billion-parameter multimodal model supporting over 140 languages and a 128K-token context window. It achieves an MMLU-Pro score of 67.5 and a LiveCodeBench score of 29.7, indicating strong performance in reasoning and coding tasks. \texttt{DeepSeek} is a distilled version of DeepSeek's R1 model, a top-performing reasoning model, fine-tuned to improve logical inference and problem-solving abilities. \texttt{Phi 4} is a 3.8-billion-parameter language model trained on high-quality web and synthetic data, significantly outperforming recent open-source models of similar size and matching the performance of models twice its size on math and coding tasks requiring complex reasoning \cite{dubey2024llama3herdmodels,mistral_nemo,gemma_2025,deepseekai2025deepseekr1incentivizingreasoningcapability,microsoft2025phi4minitechnicalreportcompact}.  We decided to focus on open-source models as we do have no control over the responses of commercial LLMs and we do not know if there are moderation layers that are not from the model itself, but business layers \cite{Spirling2023,Ollion2024}.} 

\subsection{Country Language Prompt}
\label{sec:context-prompt}

To address \textit{\textbf{RQ1}}, \textbf{\textit{RQ2}} and \textbf{\textit{RQ3}}, we used CountryHate data to prompt the models with located personas.

\paragraph{Persona} The concept of ``persona'' or ``user persona'' started in marketing and design to better understand customer behaviours and motivations. Today, user-personas are widely used for role-playing and imitating human behaviour in areas like software development, UX design, and AI research. Generally, a persona refers to a fictional character created to represent different types of users that might interact with a system, or, as \citet{zhang2024personalizationlargelanguagemodels} defines it ``an entity whose viewpoints and behaviour the simulation seeks to examine and reproduce''. This persona can be categorised in demographic persona, character persona and individualised persona \cite{chen2024personapersonalizationsurveyroleplaying}. We will focus on the first one to aggregate characteristics of demographic segments. In this regards, \citet{gupta2024personabias} propose three persona templates to guide the creation of these characters. We use \textcolor{black}{one of} their templates, adding geographic information, to prompt our models (see Table \ref{tab:personas}). We selected prompt B to examine how demographic personas affect classification outcomes compared to prompts without country information. \textcolor{black}{Future work could explore the sensibility of LLMs when prompted with the different prompts proposed by literature.}

\begin{table}[htb]
    \centering
    \begin{tabular}{r p{6.4cm}}
        \toprule
        \textbf{Id.} & \textbf{Instruction} \\ \midrule
        \textbf{A} & You are \{\textit{persona}\}. Your responses should closely mirror the knowledge and abilities of this persona. \\ 
        \textbf{B} & Adopt the identity of \{\textit{persona}\}. Answer the questions while staying in strict accordance with the nature of this identity. \\
        \textbf{C} & Take the role of \{\textit{persona}\}. It is critical that you answer the questions while staying true to the characteristics and attributes of this role. \\
        \bottomrule
    \end{tabular}
    \caption{\citet{gupta2024personabias} user personas. We used persona B in our experiments, adapting it to ``Adopt the identity of someone from \{\textit{country}\}''.}
    \label{tab:personas}
\end{table}

\paragraph{Task prompt} Our main objective is to classify hate speech, therefore, we used the following task prompt ``\textit{Determine whether a social media message is hateful or not. Respond with either a True for hate speech or False for neutral. Do not include nothing else than True or False.}''. For example, for the hate speech message ``\textit{If a song is sung by a girl there is a 94\% chance I'm going to hate it}'', for the country Afghanistan, we will use the prompt: ``\textit{Adopt the identity of someone from Afghanistan. Answer the questions while staying in strict accordance with the nature of this identity. Determine whether a social media message is hateful or not. Respond with either a True for hate speech or False for neutral. Do not include nothing else than True or False.\textless Message\textgreater If a song is sung by a girl there is a 94\% chance I'm going to hate it\textless /Message\textgreater}''.

\paragraph{Languages} To investigate whether LLMs show bias when prompted in different languages, we conducted experiments using two language settings: (1) English, the original language of the messages in the CountryHate dataset, and (2) the official language of the persona's country (we will refer to this variant as \textbf{lang}). For the second setting, we translated the original posts into the official languages of the respective persona prompt before prompting the models. \textcolor{black}{We opted to translate the data rather than use a different dataset to ensure a direct comparison between language settings—otherwise, the experiments would not be truly comparable, as they would be evaluating model behaviour on entirely different data rather than the same content presented in multiple languages.}

\subsection{Setup} CountryHate consists of \num{24132} instances, which we split into two subsets: training (\num{19306} instances) and testing (\num{4826} instances). We used the training subset for \textit{\textbf{RQ4}} (\nameref{sec:debias}), and the testing subset for studying if the LLMs exhibit a bias. We prompted each LLM \textcolor{black}{thirteen} times (12 country personas \texttt{+} 1 non-country prompt), in a zero-shot setting. This resulted in \textcolor{black}{\num{62738}} generations—\num{4826} without any context, and the remaining instances with country context\footnote{Note that the distribution of posts across countries is uneven if the country is included in CountryHate.}. This process was repeated for each language configuration (i.e. (1) English, (2) country persona language (\textbf{lang})). In total, we generated \textcolor{black}{\num{125476}} responses per model. To reduce randomness in the generation process, we set the temperature to \num{0}, top-p to \num{0.1}, top-k to \num{5}, and limited the maximum token generation to \num{256}. To summarise, our initial experiments for each model included the following variants: (1) \textbf{baseline}: no country context, (2) \textbf{country}: with country context, (3) \textbf{lang}: no country context, with posts translated into the country's language, and (4) \textbf{country lang}: both country context and posts translated into the country's language.

\paragraph{Output processing} The prompt was crafted to guide the models to output only ``True'' or ``False''. However, some variations naturally occurred (e.g., \textit{true}, \textit{yes}, \textit{vraiment} for True, and \textit{false}, \textit{no}, \textit{faux} for False). We standardised these variations. Invalid responses (e.g., ``\textit{I cannot perform this action}'') were removed. 

\section{Analysis Results}
\label{sec:results-rq1-2}
Next, we present the results for \textit{\textbf{RQ1}}, \textit{\textbf{RQ2}} and \textbf{\textit{RQ3}}, which analyse geographic and language biases in LLMs.

\paragraph{Country-specific prompts introduce bias in LLMs} To answer \textit{\textbf{RQ1}} we compare the contextualised generations with the baseline setting. We observed that the non-context variants consistently achieved higher F1 scores across all metrics for all models. Table \ref{tab:base-f1} highlights this trend, showing that \texttt{Llama 3.1} \textbf{baseline} outperforms its \textbf{country} persona counterpart (F1-macro \num{0.7428} vs. \textcolor{black}{\num{0.6269}}), and \texttt{Nemo}, \texttt{Gemma}, \texttt{DeepSeek} and \texttt{Phi 4} \textbf{baseline} similarly achieve superior results (F1-macro \num{0.8060} vs. \textcolor{black}{\num{0.7512}; \num{0.6081} vs. \num{0.5609}; \num{0.7409} vs. \num{0.5542}; and \num{0.6643} vs. \num{0.5964}, respectively}). These findings indicate that incorporating geographic context negatively impacts model behaviour.

\begin{table}[ht]
    \centering
    \renewcommand{\arraystretch}{1.3}
    \setlength{\tabcolsep}{8pt}
    \resizebox{\columnwidth}{!}{
    \begin{tabular}
        {l@{\hspace{1\tabcolsep}} rr rr}
        \toprule 
        & \multicolumn{2}{c}{\texttt{baseline}} & \multicolumn{2}{c}{\texttt{country}} \\
        \cmidrule(lr){2-3} \cmidrule(lr){4-5}
        \textbf{Model} & \multicolumn{1}{c}{\textbf{F1}} & \multicolumn{1}{c} {\textbf{F1\textsubscript{MACRO}}} & \multicolumn{1}{c}{\textbf{F1}} & \multicolumn{1}{c} {\textbf{F1\textsubscript{MACRO}}} \\
        \midrule
        \texttt{Llama 3.1} & \textbf{0.7918} & \textbf{0.7428} & \textcolor{black}{0.7401} & \textcolor{black}{0.6269} \\ \midrule
        \texttt{Nemo} & \textbf{0.8397} & \textbf{0.8060} & \textcolor{black}{0.7982} & \textcolor{black}{0.7512} \\ \midrule
        \texttt{\textcolor{black}{Gemma}} & \textcolor{black}{\textbf{0.6804}} & \textcolor{black}{\textbf{0.6081}}  & \textcolor{black}{0.6587} & \textcolor{black}{0.5609} \\ \midrule
        \texttt{\textcolor{black}{DeepSeek}} & \textcolor{black}{\textbf{0.7964}} & \textcolor{black}{\textbf{0.7409}} & \textcolor{black}{0.5944} & \textcolor{black}{0.5542} \\\midrule
        \texttt{\textcolor{black}{Phi 4}} & \textcolor{black}{\textbf{0.6903}} & \textcolor{black}{\textbf{0.6643}} & \textcolor{black}{0.6123} & \textcolor{black}{0.5964} \\
        \bottomrule
    \end{tabular}
    }
    \caption{F1 scores of Llama 3.1, Nemo, \textcolor{black}{Gemma, DeepSeek and Phi 4} base models, using the baseline (no context) and country settings.}
    \label{tab:base-f1}
\end{table}

\paragraph{Using personas' country languages reveals bias} To answer \textit{\textbf{RQ2}}, if we compare the \textbf{baseline} variant from Table \ref{tab:base-f1} and \textbf{lang} variant from Table \ref{tab:base-lang-f1}, we see that for all models \textcolor{black}{but \texttt{Gemma}} the F1 scores are lower in the \textbf{lang} setting (e.g. for \texttt{Nemo}, the \textbf{baseline} F1 score is \num{0.8397}, and for \textbf{lang} is \num{0.8261}). Moreover, for \textbf{\textit{RQ3}} we see that the use of personas' country languages further exacerbates the decline in model performance seen with geographic context alone. For example, \texttt{Llama 3.1} \textbf{country lang} achieves an F1-macro score of \textcolor{black}{\num{0.5993}}, a notable drop from the \textbf{country} variant's \textcolor{black}{\num{0.6269}} in Table \ref{tab:base-f1}. Similarly, \texttt{Nemo} \textbf{country lang} F1-macro score of \textcolor{black}{\num{0.7182}} is lower than the \textbf{country} variant’s \textcolor{black}{\num{0.7512}}. \textcolor{black}{And the same behaviour is seen in the rest of the models between the \textbf{country} and \textbf{country lang} variants.} This comparison underscores that incorporating both geographic context and the country's official language amplifies biases, resulting in a more pronounced reduction in overall model performance. Moreover, we noticed that for \texttt{Llama 3.1}, \textcolor{black}{\texttt{Gemma} and \texttt{DeepSeek}}, the \textbf{country lang} variant yielded a high number of invalid generations \textcolor{black}{(\textgreater \num{7000})}, suggesting that incorporating both geographic and language context may have introduced complexities or ambiguities that the model struggled to handle effectively.

\begin{table}[ht]
    \centering
    \renewcommand{\arraystretch}{1.3}
    \setlength{\tabcolsep}{8pt}
    \resizebox{\columnwidth}{!}{
    \begin{tabular}
        {l@{\hspace{1\tabcolsep}} rr rr}
        \toprule 
        & \multicolumn{2}{c}{\texttt{lang}} & \multicolumn{2}{c}{\texttt{country lang}} \\
        \cmidrule(lr){2-3} \cmidrule(lr){4-5}
        \textbf{Model} & \multicolumn{1}{c}{\textbf{F1}} & \multicolumn{1}{c} {\textbf{F1\textsubscript{MACRO}}} & \multicolumn{1}{c}{\textbf{F1}} & \multicolumn{1}{c} {\textbf{F1\textsubscript{MACRO}}} \\
        \midrule
        \texttt{Llama 3.1} & \textbf{0.7880} & \textbf{0.7389} & \textcolor{black}{0.7223} & \textcolor{black}{0.5993} \\ \midrule
        \texttt{Nemo} & \textbf{0.8261} & \textbf{0.7910} & \textcolor{black}{0.7634} & \textcolor{black}{0.7182} \\ \midrule
        \texttt{\textcolor{black}{Gemma}} & \textcolor{black}{\textbf{0.6945}} & \textcolor{black}{\textbf{0.6200}}  & \textcolor{black}{0.5840} & \textcolor{black}{0.4945} \\ \midrule
        \texttt{\textcolor{black}{DeepSeek}} & \textcolor{black}{\textbf{0.7517}} & \textcolor{black}{\textbf{0.6926}} & \textcolor{black}{0.5821} & \textcolor{black}{0.5347} \\\midrule
        \texttt{\textcolor{black}{Phi 4}} & \textcolor{black}{\textbf{0.6418}} & \textcolor{black}{\textbf{0.6207}} & \textcolor{black}{0.5341} & \textcolor{black}{0.5279} \\
        \bottomrule
    \end{tabular}
    }
    \caption{F1 scores of Llama 3.1, Nemo \textcolor{black}{Gemma, DeepSeek and Phi 4}, on the testing subset.}
    \label{tab:base-lang-f1}
\end{table}

\paragraph{\textcolor{black}{Llama 3.1 is most affected by context}} The results in Table \ref{tab:base-fnr} reveal that \texttt{Llama 3.1} exhibits \textcolor{black}{the largest increase in False Negative Rates (FNR) when moving from the \textbf{baseline} variant (\num{32.37}\%) to the \textbf{country} and \textbf{country-lang} variants (over \num{71}\%). In contrast, \texttt{Nemo}'s FNR rises more moderately, from \num{16.95}\% to \num{30.66}\% and \num{26.61}\%, respectively, suggesting a lower sensitivity to geographic context. \texttt{Gemma} also shows a notable increase, with FNRs climbing from \num{54.11}\% to \num{81.43}\%. \texttt{DeepSeek} yielded an slightly lower FNR for the country variant compared to the baseline and lang ones, but the performance in terms of F1-score is much lower when prompted with country context. \texttt{Phi 4}, despite being a relatively weak performer in terms of F1-scores, maintains low FNRs across all settings, indicating minimal fluctuation due to context. These patterns highlight that while all models exhibit some level of bias, \texttt{Llama 3.1} is the most affected by contextual changes.}

\begin{table}[ht]
    \centering
    \renewcommand{\arraystretch}{1.3}
    \setlength{\tabcolsep}{8pt}
    \resizebox{\columnwidth}{!}{
    \begin{tabular}
        {l@{\hspace{1\tabcolsep}} rrrr}
        \toprule
        \textbf{Model} & \makecell[c]{\texttt{baseline}} & \makecell[c]{\texttt{country}} & \makecell[c]{\texttt{lang}} & \makecell[c]{\texttt{country} \\ \texttt{lang}} \\
        \midrule
        \texttt{Llama 3.1} & 32.37\%  & \textcolor{black}{71.42\%} & \textbf{32.22\%} & \textcolor{black}{71.20\%} \\ 
        \midrule
        \texttt{Nemo} & \textbf{16.95\%} & \textcolor{black}{30.66\%} & 17.48\% & \textcolor{black}{26.61\%} \\
        \midrule
        \texttt{\textcolor{black}{Gemma}} & \textcolor{black}{\textbf{54.11\%}} & \textcolor{black}{75.13\%} & \textcolor{black}{56.26\%} & \textcolor{black}{81.43\%} \\
        \midrule
        \texttt{\textcolor{black}{DeepSeek}} & \textcolor{black}{42.36\%} & \textcolor{black}{\textbf{31.79\%}} & \textcolor{black}{41.13\%} & \textcolor{black}{40.70\%} \\
        \midrule
        \texttt{\textcolor{black}{Phi 4}} & \textcolor{black}{7.67\%} & \textcolor{black}{\textbf{6.06\%}} & \textcolor{black}{7.60\%} & \textcolor{black}{8.41\%} \\
        \bottomrule
    \end{tabular}
    }
    \caption{False Negative Rates (FNR) of Llama 3.1, Nemo, \textcolor{black}{Gemma, DeepSeek and Phi 4}, on the testing subset.}
    \label{tab:base-fnr}
\end{table}

\paragraph{Llama 3.1 exhibit country-specific bias} As \texttt{Llama 3.1} showed \textcolor{black}{the} higher bias \textcolor{black}{within all models}, we examined the differences across countries for \texttt{Llama 3.1}. Countries such as Brunei, \textcolor{black}{Cuba, Russia}, and Saudi Arabia exhibited a significantly higher FNR—approximately \textcolor{black}{140}\% higher than the baseline (\texttt{Llama 3.1} \textbf{baseline}). This indicates that the model was more likely to miss classifying hateful messages as such in these countries. In contrast, when the country context was United Kingdom (UK) or United States (US), the false negative rate was notably lower—around 80\% of the baseline FNR. Nigeria and Uganda \textcolor{black}{followed with the next lowest rates, though their values remained higher than those of the UK and US (approx. 10\% more).} We performed a chi-squared test (${\mathcal{X}^2}$) at ${p > 0.01}$ to determine whether there is a statistically significant difference between our baseline predictions and the contextualised predictions. As we can see in Table \ref{tab:fnr_significance} all the countries reject the null hypothesis (${p < 0.01}$): adding country context significantly affects the LLM's classification.

\setlength{\tabcolsep}{0.5em}
\begin{table}[ht]
    \centering
    \begin{tabular}{l rrr}
        \toprule
        \textbf{Country} & \textbf{FN} & \textbf{FNR} & \textbf{p-value} \\ \midrule
        \texttt{baseline} & 426 & 32.37\% & -- \\ \midrule
        Afghanistan & 858 & {66.67\%} & $2.78e\textsuperscript{-65}$ \\
        \textcolor{black}{Belarus} & \textcolor{black}{821} & \textcolor{black}{70.23\%} & \textcolor{black}{$2.21e\textsuperscript{-63}$} \\
        Brunei & 965 & {78.58\%} & $4.17e\textsuperscript{-114}$ \\
        \textcolor{black}{China} & \textcolor{black}{815} & \textcolor{black}{64.33\%} & \textcolor{black}{$1.98e\textsuperscript{-57}$} \\
        \textcolor{black}{Cuba} & \textcolor{black}{891} & \textcolor{black}{79.77\%} & \textcolor{black}{$2.16e\textsuperscript{-99}$} \\
        \textcolor{black}{Nicaragua} & \textcolor{black}{936} & \textcolor{black}{76.47\%} & \textcolor{black}{$3.88e\textsuperscript{-93}$} \\
        Nigeria & 817 & 62.08\% & $2.55e\textsuperscript{-60}$ \\
        \textcolor{black}{North Korea} & \textcolor{black}{517} & \textcolor{black}{91.67\%} & \textcolor{black}{$3.98e\textsuperscript{-67}$} \\
        Qatar & 897 & {72.46\%} & $4.60e\textsuperscript{-90}$ \\
        \textcolor{black}{Russia} & \textcolor{black}{958} & \textcolor{black}{78.91\%} & \textcolor{black}{$3.21e\textsuperscript{-102}$} \\
        Saudi Arabia & 845 & {77.52\%} & $6.63e\textsuperscript{-100}$ \\
        Uganda & 815 & 62.02\% & $6.68e\textsuperscript{-56}$ \\ \midrule
        Australia & 78 & {69.64\%} & $1.51e\textsuperscript{-6}$ \\
        UK & 27 & 54.00\% & $6.04e\textsuperscript{-82}$ \\
        US & 554 & 59.76\% & $1.53e\textsuperscript{-4}$ \\
        \bottomrule
    \end{tabular}
    \caption{False Negatives (FN), False Negative Rate per Hate Cases (FNR), and p-values for the selected countries and baseline, for \texttt{Llama 3.1}.}
    \label{tab:fnr_significance}
\end{table}

\textcolor{black}{As \texttt{Nemo} was our best performing model, we decided to examine the country-specific bias.} For \textcolor{black}{this model}, a different pattern emerges. Our target countries do not show higher false negative rates, but Australia stands out with elevated rates, although these are lower than those of \texttt{Llama 3.1} (\num{46.67}\% vs. \num{69.00}\%). This higher false negative rate in Australia matches the results from \texttt{Llama 3.1} and aligns with research indicating that there are strong biases against immigrants and refugees in Australia \cite{Schweitzer2005}. Because of this, when using a persona prompt with Australia, the model might not recognise hate speech against refugees or immigrants, as it has learned these prejudices about the Australian population. For the other countries, the false negative rates stay about the same.

\textcolor{black}{In summary, prompting models with persona prompts leads to a performance drop, which varies depending on the specific country, not only in the incorporation of the prompt. While this work focuses on geographic and language bias, similar patterns have been observed in how LLMs represent emotions related to religion \cite{plaza-del-arco-etal-2024-divine}. Future research could further explore other bias in LLMs, particularly in the context of hate speech identification.}

\section{Mitigate Bias}
\label{sec:debias}

Based on our findings in \textit{\textbf{RQ1}}, \textit{\textbf{RQ2}} and \textbf{\textit{RQ3}}, which revealed demographic bias in the selected LLMs, we propose \textit{debias tuning} as a strategy to reduce this bias. The key idea of debias tuning is to align hate speech classification outcomes with contextual information and without it, while reducing the impact of biases in the context \cite{dong2024disclosuremitigationgenderbias}. Prior research has shown the effectiveness of prompting techniques for mitigating biases such as reducing gender bias, limiting the generation of hate speech or even preventing a model from performing certain tasks \cite{2023dwivedi,piot2024decodinghateexploringlanguage}. In this work, we present a debias tuning method to reduce inconsistencies between country-context and non-context predictions, ensuring more consistent responses across locations.

\paragraph{Setup} We fine-tuned \texttt{Llama 3.1}, \texttt{Nemo} \textcolor{black}{and \texttt{Phi 4}} models using the training subset. \textcolor{black}{This model selection allowed us to assess the generalizability of the approach while keeping the computational cost manageable.} For our fine-tuning, we provide the models with the persona prompt, the social media text and the gold label. \textcolor{black}{In order to study the generability of the models, on both country and language dimensions, we decided to perform the finetuning with only four countries from our candidates list. These are Afghanistan (Persian), Brunei (Malay), Qatar (Arabic) and Saudi Arabia (Arabic)}.  We fine-tuned the models in two language settings: (1) English (\texttt{debias tuning}), and (2) the official languages of the personas' countries (\texttt{multilingual debias tuning}).

\paragraph{Custom loss} To introduce our proposed debias, we have implemented a custom loss. Our loss applies a consistency penalty if either the prediction is invalid (i.e., not in {True, False}), or if the non-context prediction matches the gold label, but the context prediction does not. Let ${x}$ be the input text, ${x_c}$ the input text with country context text and ${y}$ the gold label. We use the cross-entropy function as our loss function, where ${\mathcal{L}_{\text{class}} = CrossEntropy(logits(x), y))}$ is the classification loss for the non-country input, and ${\mathcal{L}_{\text{class}}^c = CrossEntropy(logits(x_c), y))}$ is the loss for the country-context text. The average classification loss is computed as:

\begin{equation}
\mathcal{L}_{\text{avg\_class}} = \frac{\mathcal{L}_{\text{class}} + \mathcal{L}_{\text{class}}^c}{2}
\label{eq:loss}
\end{equation}

Then, to compute the loss with the consistency penalty, we include Equation \ref{eq:loss} as:

\begin{equation}
\mathcal{L}_{\text{loss}} = \mathcal{L}_{\text{avg\_class}} + \alpha \cdot \mathcal{L}_{\text{avg\_class}}
\label{eq:loss_consistency}
\end{equation}

Note that in Equation \ref{eq:loss_consistency}, ${\alpha}$ is $> 0$ if ${\hat{y} = y \text{ and } \hat{y}_c \neq y}$ or if ${\hat{y} = {None} \text{ or } \hat{y}_c = {None}}$. Otherwise, this term is ${0}$. 

\paragraph{Debias fine-tuning} We fine-tuned \texttt{Meta-Llama-3.1- 8B-Instruct-bnb-4bit}, \texttt{Mistral-Nemo-Ins- truct-2407-bnb-4bit} \textcolor{black}{and \texttt{Phi-4-mini- instruct-unsloth-bnb-4bit}} from unsloth using 4-bit precision for memory efficiency. Models are initialised with a sequence length of \num{256} tokens and fine-tuned with LoRA for one epoch, using an effective batch size of \num{16} (batch size \num{4}, gradient accumulation \num{4}). We employ the AdamW optimizer in \num{8}-bit precision with a learning rate of $2e\textsuperscript{-6}$, weight decay of \num{0.01}, and gradient clipping at \num{0.3}. Training includes a \num{3}\% warm-up and linear decay for the learning rate. 

With our models fine-tuned (\texttt{debias-llama}, \texttt{debias-nemo}, \textcolor{black}{\texttt{debias-phi},} \texttt{debias-llama-lang}, \texttt{debias-nemo-lang}, \textcolor{black}{and \texttt{debias-phi-lang}}), we proceed to evaluate them. We used the testing subset and the same persona prompt, the \textcolor{black}{twelve} selected countries and the original author's country. Next we present the results.

\section{Debias Tuning Results}

\paragraph{Debias tuning \textcolor{black}{outperforms base models}} Comparing the F1 scores from the debias models to the base models reveals that for the \textbf{baseline} all \texttt{debias-llama}, \texttt{debias-nemo} \textcolor{black}{and \texttt{debias-phi}} achieve higher F1 scores (see Table \ref{tab:debias-f1}, row baseline). With \textbf{country} context, \texttt{debias-llama} notably increases the F1 scores, specially the F1-macro (from \textcolor{black}{\num{0.6269}} to \textcolor{black}{\num{0.7169}}). \texttt{debias-nemo} \textcolor{black}{and \texttt{debias-phi}} models achieve similar but slightly lower F1 scores improvements, compared to its base version. This outcome may be due to the fact that \texttt{Nemo} \textcolor{black}{and \texttt{Phi 4}} did not exhibit as much bias in its base version, indicating that its initial robustness may limit the extent of improvement possible through debiasing, \textcolor{black}{but still getting improved results}.

\begin{table*}
    \centering
    \renewcommand{\arraystretch}{1.4}
    \setlength{\tabcolsep}{5pt}
    \resizebox{\textwidth}{!}{
    \begin{tabular}{ll | ccc | ccc | ccc}
        \toprule
        \thead{\textbf{Variant}} & \thead{\textbf{Metric}} & \thead{\texttt{Llama 3.1}} & \thead{\texttt{debias} \\ \texttt{llama}} & \thead{\texttt{debias} \\ \texttt{llama-lang}} & \thead{\texttt{Nemo}} & \thead{\texttt{debias} \\ \texttt{nemo}} & \thead{\texttt{debias} \\ \texttt{nemo-lang}}  & \thead{\texttt{Phi 4}} & \thead{\texttt{debias} \\ \texttt{phi}} & \thead{\texttt{debias} \\ \texttt{phi-lang}} \\
        \midrule
        \multirow{2}{*}{\texttt{baseline}} & F1 & 0.7918 & 0.8001 & \textbf{0.8142} & 0.8397 & \textbf{0.8518} & 0.8413 & \textcolor{black}{0.6903} & \textcolor{black}{\textbf{0.6905}} & \textcolor{black}{\textbf{0.6905}} \\
         & F1\textsubscript{MACRO} & 0.7428 & 0.7655 & \textbf{0.7765} & 0.8060 & \textbf{0.8155} & 0.7906 & \textcolor{black}{0.6643} & \textcolor{black}{\textbf{0.6645}} & \textcolor{black}{\textbf{0.6645}} \\
        \midrule
        \multirow{2}{*}{\texttt{country}} & F1 & \textcolor{black}{0.7401} & \textcolor{black}{0.7792} & \textcolor{black}{\textbf{0.7937}} & \textcolor{black}{0.7982} & \textbf{\textcolor{black}{0.8009}} & \textcolor{black}{0.7929} & \textcolor{black}{0.6123} & \textcolor{black}{\textbf{0.6154}} & \textcolor{black}{0.6122} \\
         & F1\textsubscript{MACRO} & \textcolor{black}{0.6269} & \textcolor{black}{0.7169} & \textbf{\textcolor{black}{0.7323}} & \textcolor{black}{0.7512} & \textbf{\textcolor{black}{0.7523}} & \textcolor{black}{0.7264} & \textcolor{black}{0.5964} & \textcolor{black}{\textbf{0.5988}} & \textcolor{black}{0.5962} \\
        \midrule
        \multirow{2}{*}{\texttt{lang}} & F1 & 0.7880 & 0.8098 & \textbf{0.8184} & 0.8261 & \textbf{0.8431} & 0.8416 & \textcolor{black}{0.6418} & \textcolor{black}{0.6441} & \textcolor{black}{\textbf{0.6493}} \\
         & F1\textsubscript{MACRO} & 0.7389 & 0.7661 & \textbf{0.7814} & 0.7910 & \textbf{0.8057} & 0.7907 & \textcolor{black}{0.6207} & \textcolor{black}{0.6229} & \textcolor{black}{\textbf{0.6269}} \\
        \midrule
        \multirow{2}{*}{\texttt{country lang}} & F1 & \textcolor{black}{0.7223} & \textcolor{black}{0.7679} & \textbf{\textcolor{black}{0.7972}} & \textcolor{black}{0.7634} & \textcolor{black}{0.7831} & \textcolor{black}{\textbf{0.7894}} & \textcolor{black}{\textbf{0.5341}} & \textcolor{black}{0.5298} & \textcolor{black}{0.5316} \\
         & F1\textsubscript{MACRO} & \textcolor{black}{0.5993} & \textcolor{black}{0.7182} & \textbf{\textcolor{black}{0.7411}} & \textcolor{black}{0.7182} & \textcolor{black}{\textbf{0.7344}} & \textcolor{black}{0.7242} & \textcolor{black}{\textbf{0.5279}} & \textcolor{black}{0.5244} & \textcolor{black}{0.5260} \\ 
        \bottomrule
    \end{tabular}
    }
    \caption{F1 scores of base models and debias models, with the non-context and country variants, on the testing subset.}
    \label{tab:debias-f1}
\end{table*}

\paragraph{Mixed outcomes in multilingual debiasing} The results in Table \ref{tab:debias-f1}, rows baseline and country, show that the \texttt{debias-llama-lang} model consistently outperforms its counterparts across both the \textbf{baseline} and \textbf{country} variant. In the \textbf{baseline} setting, \texttt{debias-llama-lang} achieves an F1-micro score of \num{0.8142}, surpassing  \texttt{Llama 3.1} (\num{0.7918}) and the English-tuned-only \texttt{debias-llama} (\num{0.8001}). Notably, in the \textbf{country} variant, \texttt{debias-llama-lang} records an F1-micro of \textcolor{black}{\num{0.7937}, getting closer to} the \textbf{baseline} (\num{0.8142}). This is a significant improvement compared to the original \texttt{Llama 3.1}, where the \textbf{country} variant (\textcolor{black}{\num{0.7401}}) was \textcolor{black}{\num{0.0538}} points lower than the \textbf{baseline} version (\num{0.7918}). This reduction in the performance gap highlights that the classification bias introduced by geographic context has been effectively mitigated in the \texttt{Llama} debiased multilingual model, allowing for consistent performance across contexts. However, the results show that the \texttt{debias-nemo-lang} model did not perform as well as the English-only \texttt{debias-nemo} version or the base model. For the \textbf{baseline}, \texttt{debias-nemo} achieved higher F1 (\num{0.8518} vs. \num{0.8413}) and F1-macro (\num{0.8155} vs. \num{0.7906}) scores. Additionally, in the country setting, \texttt{debias-nemo} outperformed \texttt{debias-nemo-lang} in both metrics (F1: \textcolor{black}{\num{0.8009}} vs. \textcolor{black}{\num{0.7929}}, F1-macro: \textcolor{black}{\num{0.7523}} vs. \textcolor{black}{\num{0.7264}}), \textcolor{black}{but both debias methods performed better than their base version}.  \textcolor{black}{For \texttt{Phi 4}, we see a similar behaviour as for \texttt{Nemo}, with the \texttt{debias-phi} method yielding the best results for the \textbf{baseline} and \textbf{country} variants.} These findings suggest that multilingual debiasing is highly effective in reducing bias for models like \texttt{Llama}, which exhibited significant bias with location context. However, in the \texttt{Nemo} \textcolor{black}{and \texttt{Phi}} variants, where the initial bias was less pronounced, the improvements were not as substantial.

\paragraph{Debias tuning reduces language bias} Table \ref{tab:debias-f1}, rows lang and country lang, demonstrates that debias tuning effectively reduces language bias, improving F1 scores across all metrics and settings for both \texttt{Llama 3.1} and \texttt{Nemo}. For \texttt{Llama 3.1}, the debiased variants show marked improvements, with the \textbf{country lang} F1-macro score increasing from \textcolor{black}{\num{0.5993}} to \textcolor{black}{\num{0.7182}} and the \textbf{lang} score rising from \num{0.7389} to \num{0.7661}. \texttt{Nemo} also benefits from debiasing, albeit to a lesser extent. The \texttt{debias-nemo} \textbf{lang} F1-macro improving from \num{0.7910} to \num{0.8057} and the \textbf{country lang} variant seeing a smaller increase from \textcolor{black}{\num{0.7182}} to \textcolor{black}{\num{0.7344}}. \textcolor{black}{For \texttt{Phi}, multilingual learning was less effective compared to the other models, with improvements observed only in the \textbf{lang} setting.} These results indicate that debias tuning improves the overall performance, with particularly strong gains for Llama, in multilingual contextual settings.

\begin{figure*}[ht]
    \centering
    \includegraphics[scale=0.21]{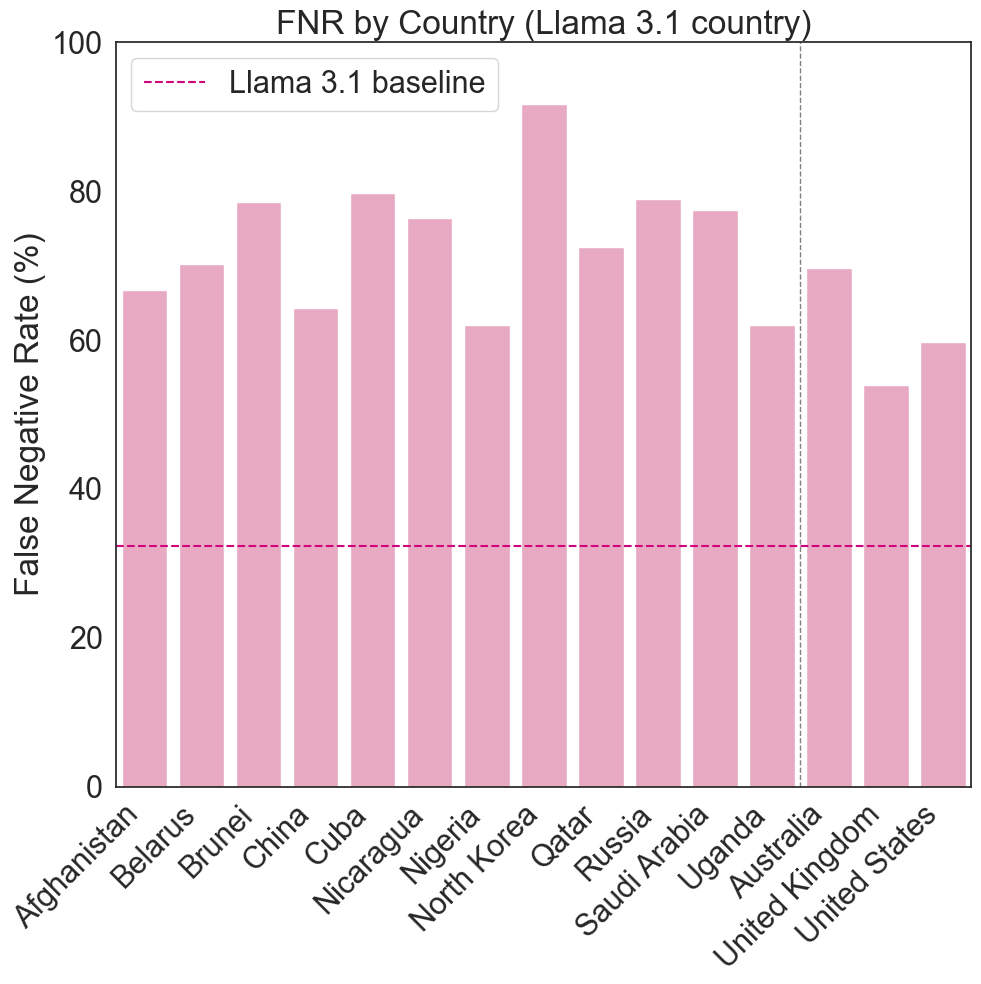}
    \includegraphics[scale=0.21]{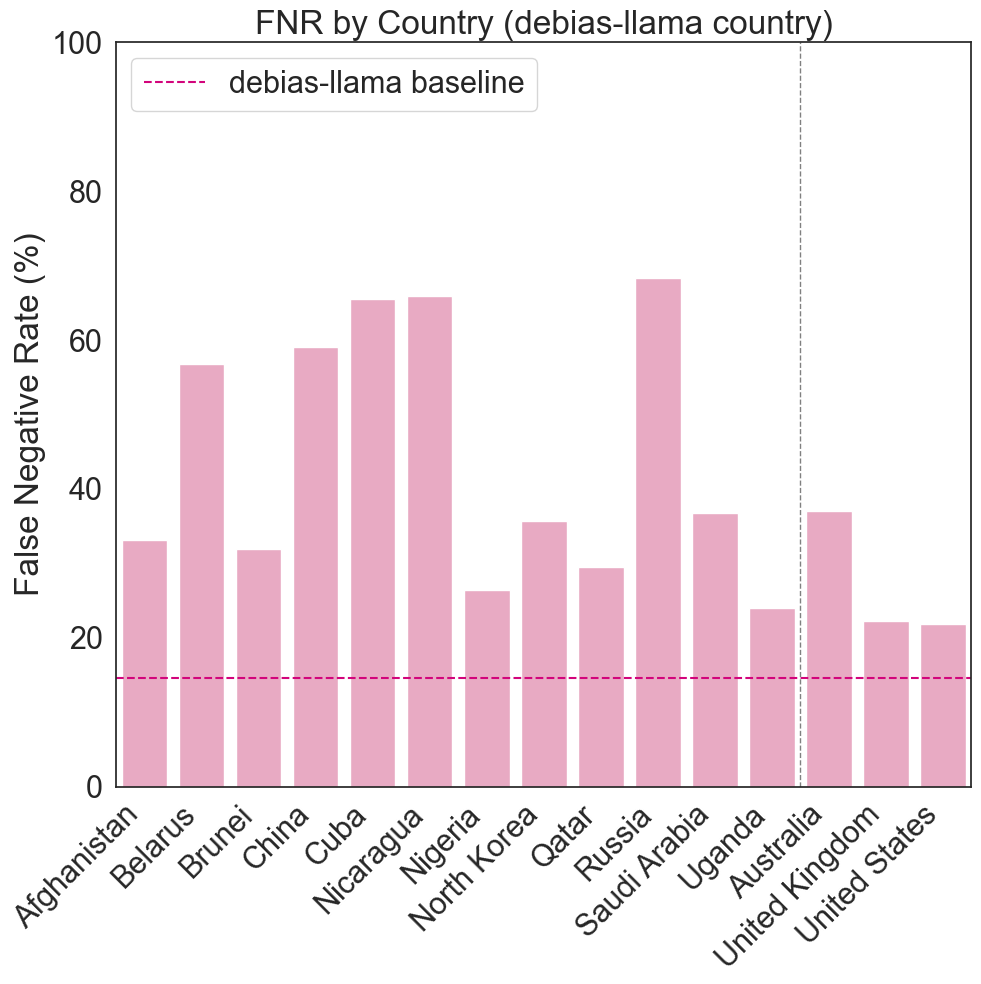}
    \includegraphics[scale=0.21]{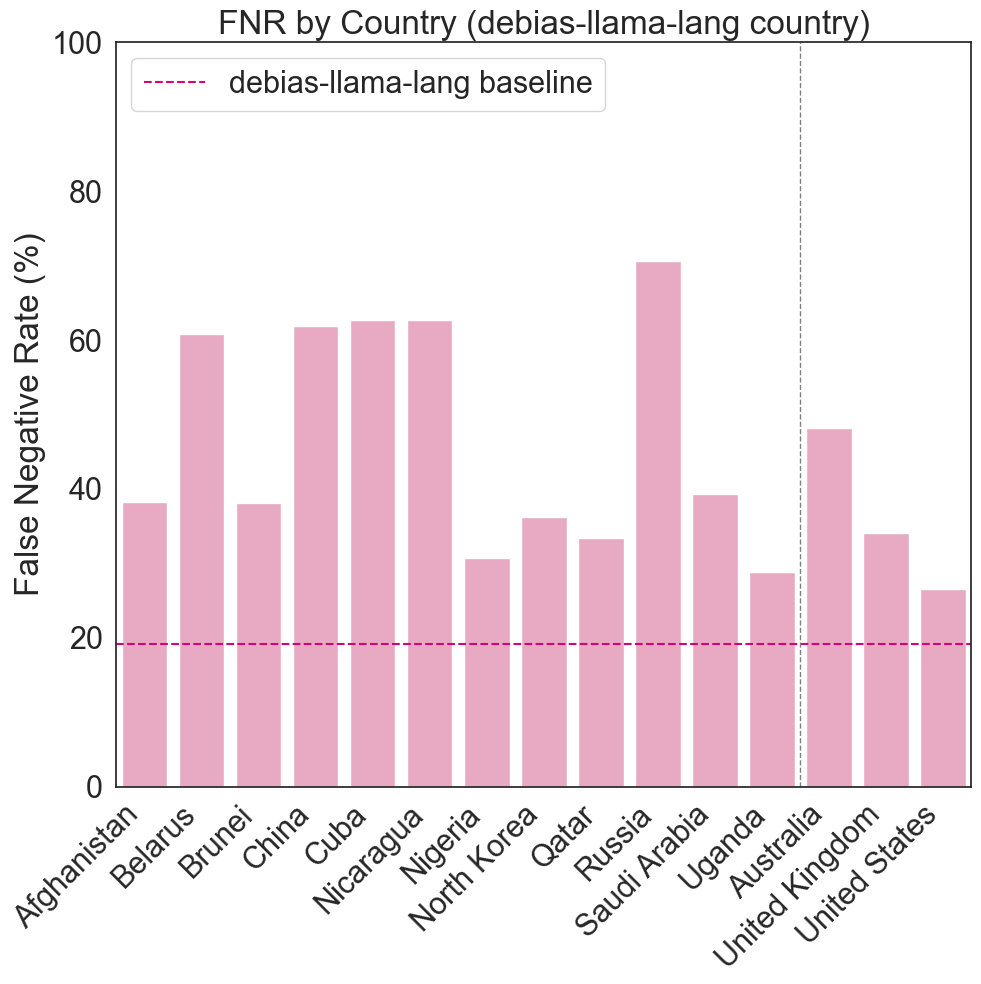}
    \includegraphics[scale=0.21]{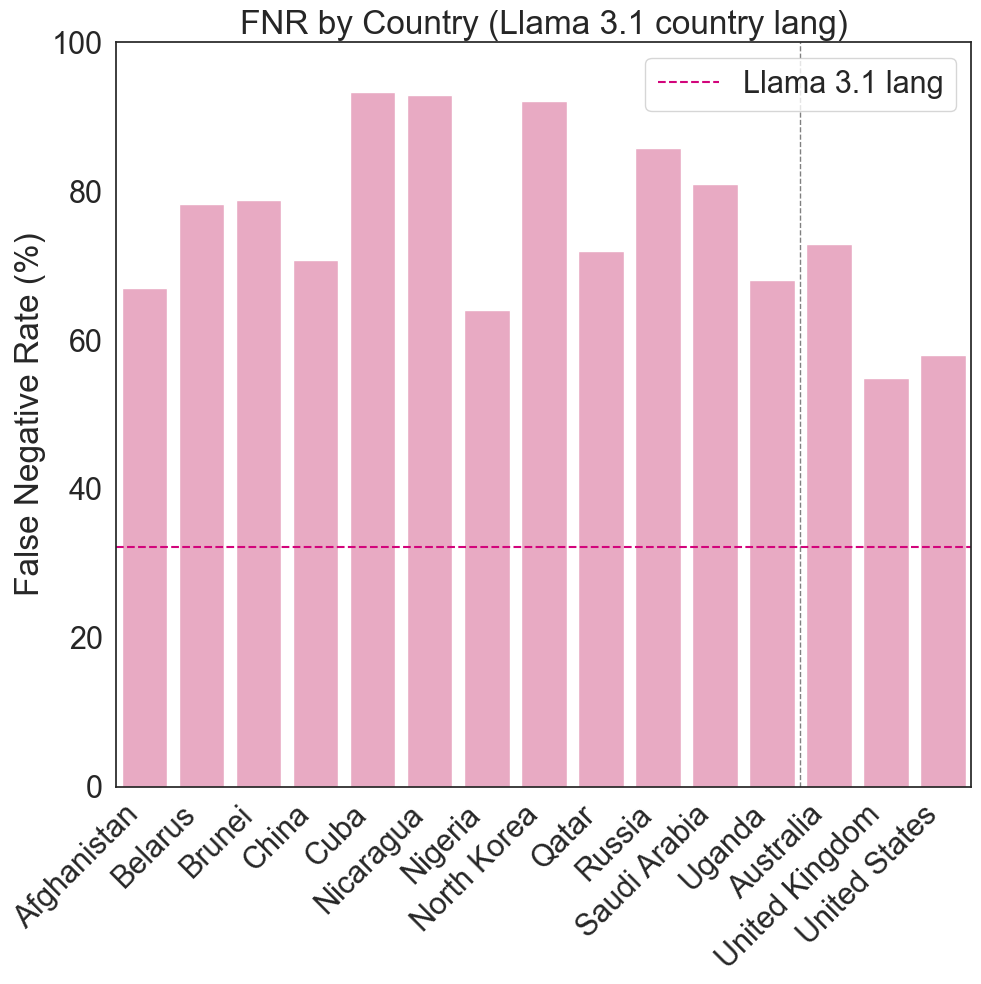}
    \includegraphics[scale=0.21]{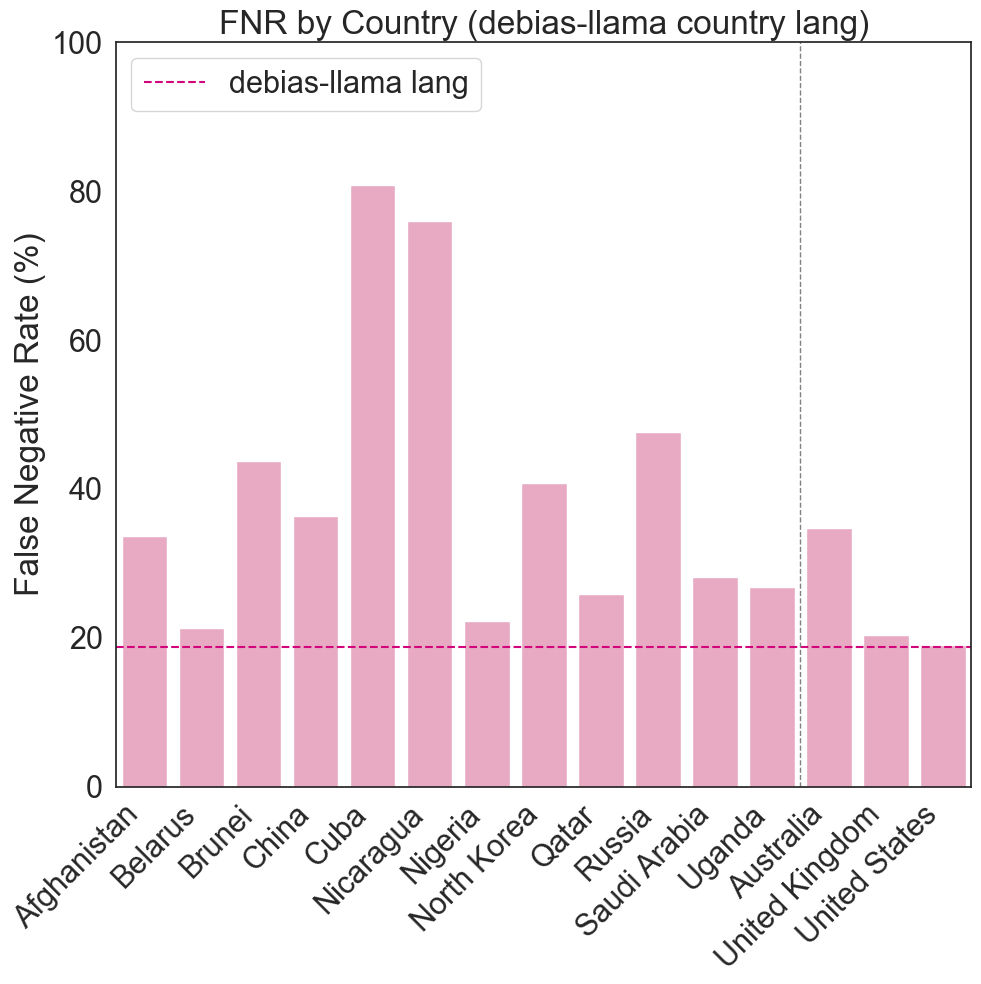}
    \includegraphics[scale=0.21]{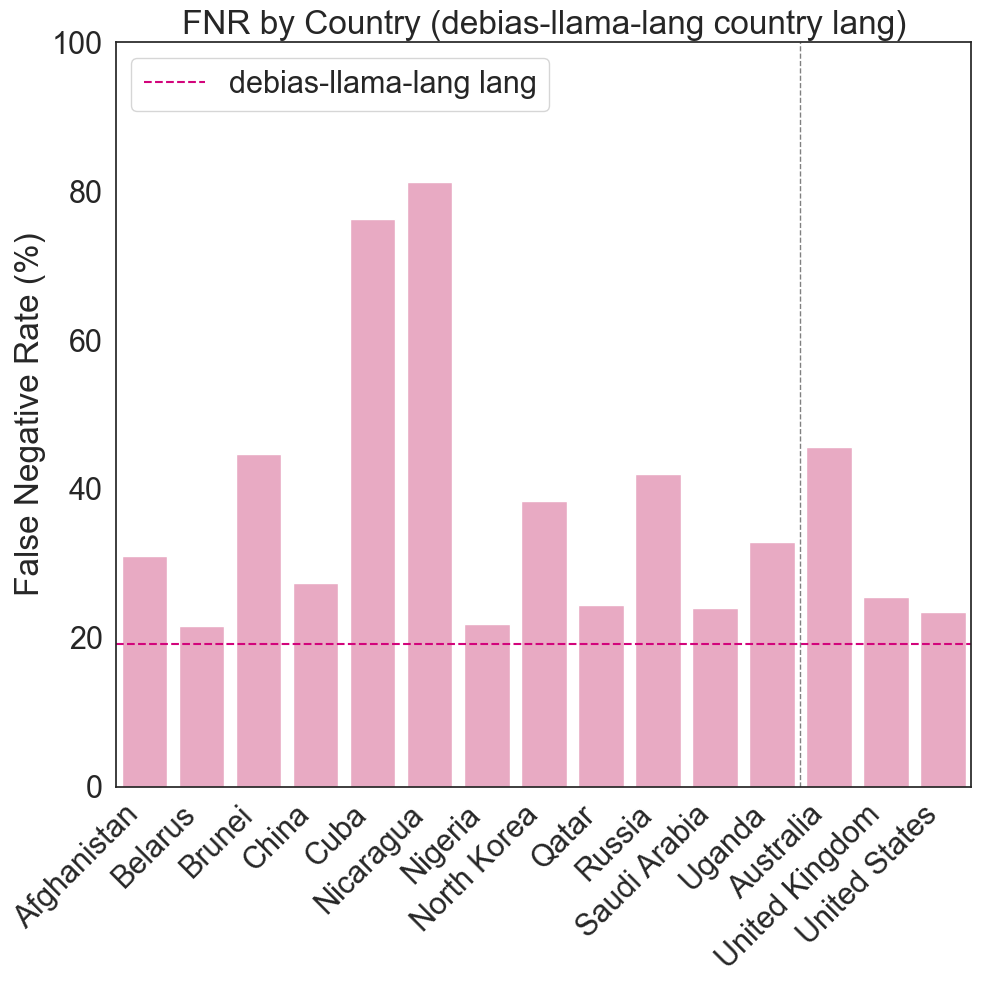}
    \caption{(Up): FNR across countries for (1) {Llama 3.1}, (2) {debias-llama} and (3) {Debias Llama Lang}, for English. (Down): FNR across countries for (4) {Llama 3.1}, (5) {debias-llama} and (6) {Debias Llama Lang}, for persona's country languages (lang).}
    \label{fig:llama-fnr}
\end{figure*}

\paragraph{Multilingual debias tuning excels in diverse language inference} Table \ref{tab:debias-f1}, rows lang and country lang, highlight the strong performance of the \texttt{debias-llama-lang} model, which consistently outperforms other Llama variants. In the \textbf{lang} setting, \texttt{debias-llama-lang} achieves an F1 score of \num{0.8184}, surpassing the original \texttt{Llama 3.1} (\num{0.7880}) and the English-only \texttt{debias-llama} (\num{0.8098}). Similarly, in the \textbf{country lang} variant, it records an F1 score of \textcolor{black}{\num{0.7972}}, significantly outperforming its counterparts. These results brings closer F1 scores for the non-context and context variants, showing that our proposed models mitigated biases introduced by contextual, personalised prompts. For the \texttt{debias-nemo-lang} model, the \textbf{country lang} variant shows improved F1 scores, increasing from \textcolor{black}{\num{0.7634}} to \textcolor{black}{\num{0.7894}}. However, in the \textbf{lang} setting, the F1 and F1-macro scores are slightly higher for \texttt{debias-nemo} (F1 \num{0.8431}, F1-macro \num{0.8057}) compared to \texttt{debias-nemo-lang} (F1 \num{0.8416}, F1-macro \num{0.7907}). \textcolor{black}{\texttt{debias-phi-lang} achieves the highest F1 score (0.6493) in the \textbf{lang} setting, while in the \textbf{country lang} setting, the base model performs best. However, the F1 scores across all three variants are similarly low and close to each other.} These findings confirm that multilingual debiasing not only reduces bias effectively but also improves model performance in multilingual contexts, making it the most reliable approach among the tested variants. \textcolor{black}{These results also underscore the importance of choosing a strong LLM, as \texttt{Phi} shows limited performance and smaller gains}. Moreover, although \texttt{debias-llama-lang} yields the best improvement, \texttt{debias-nemo-lang} still performs better in its \textbf{baseline} setting. However, in the \textbf{country lang} variant, originally, \texttt{Nemo} outperformed \texttt{Llama 3.1}, but now the debias models for Llama outperform Nemo's, showing the gains in multilingualism for this model.

\paragraph{Debias tuning reduces country-specific bias} We analysed the evolution of FNRs across the base and debias models. As we can see in Figure \ref{fig:llama-fnr}, upper row, debias tuning significantly reduces country-specific bias in \texttt{Llama 3.1} by lowering FNRs compared to baseline configurations. In the \texttt{Llama 3.1} model, countries like Afghanistan (66.67\%), Brunei (78.58\%), \textcolor{black}{North Korea (91.67\%)} and Saudi Arabia (77.52\%) had high FNRs, while the United States (59.76\%) and the United Kingdom (54.00\%) performed moderately better. \texttt{debias-llama} reduced FNRs in high-bias regions, such as Afghanistan (33.10\%), Brunei (31.94\%) \textcolor{black}{and North Korea (35.65\%)}, achieving more equitable detection. \textcolor{black}{Other countries such as Cuba, Nicaragua or Russia, show improved results (new FNRs 65.58\%, 65.94\% and 68.30\%, old FNRs 93.23\%, 92.95\% and 85.77\% respectively) but to a smaller extent.} The \texttt{debias-llama-lang} model maintained the results but showed slightly higher FNRs in some regions, like Afghanistan (38.20\%), Saudi Arabia (39.31\%) \textcolor{black}{and Russia (70.59\%)}. Table \ref{tab:significance_results} further illustrates the effectiveness of debias tuning. We performed a chi-square test (${\mathcal{X}^2}$) at ${p > 0.01}$ to evaluate the significance of false negatives across the countries compared to the United Kingdom (the country with the lowest FNR). The p-values indicate that while the \texttt{Llama 3.1} model shows significant differences across all our target countries, the \texttt{debias-llama} model eliminates significant differences \textcolor{black}{in many of the countries, and raised the p-values in the rest of them, reflecting reduced disparity}. This reduction in significance indicates that the debiasing models successfully mitigate country-specific bias, creating a more equitable performance across the targeted countries. \textcolor{black}{Nonetheless, to improve generalisation and robustness, we recommend extending the debiasing process to include a broader set of countries.}

\begin{table}[ht]
    \centering
    \begin{tabular}{@{}lccccc@{}}
        \toprule
        \multicolumn{1}{c}{} & \multicolumn{2}{c}{\texttt{Llama 3.1}} & \multicolumn{2}{c}{\texttt{debias-llama}} \\ 
        \cmidrule(lr){2-3} \cmidrule(lr){4-5} 
        \textbf{Country} & \textbf{P-value} & \textbf{Sig.} & \textbf{P-value} & \textbf{Sig.} \\ 
        \midrule
         Afghanistan     & $5.38e\textsuperscript{-5}$    & Yes   & \textcolor{black}{0.0084}  & No \\
         \textcolor{black}{Belarus}         & \textcolor{black}{$2.21e\textsuperscript{-64}$}    & \textcolor{black}{Yes}   & \textcolor{black}{$1.08e\textsuperscript{-6}$}  & \textcolor{black}{Yes} \\
         Brunei          & $2.26e\textsuperscript{-7}$    & Yes   & \textcolor{black}{0.0096}  & No \\
         \textcolor{black}{China}           & \textcolor{black}{$1.98e\textsuperscript{-57}$}    & \textcolor{black}{Yes}   & \textcolor{black}{$1.56e\textsuperscript{-6}$}  & \textcolor{black}{Yes} \\
         \textcolor{black}{Cuba}            & \textcolor{black}{$2.16e\textsuperscript{-99}$}    & \textcolor{black}{Yes}   & \textcolor{black}{$2.48e\textsuperscript{-6}$}  & \textcolor{black}{Yes} \\
         \textcolor{black}{Nicaragua}       & \textcolor{black}{$3.88e\textsuperscript{-93}$}    & \textcolor{black}{Yes}   & \textcolor{black}{$1.89e\textsuperscript{-6}$}  & \textcolor{black}{Yes} \\
         Nigeria         & 0.0003                  & Yes   & \textcolor{black}{0.0589}  & No \\
         \textcolor{black}{North Korea}     & \textcolor{black}{$3.98e\textsuperscript{-67}$}    & \textcolor{black}{Yes}   & \textcolor{black}{$1.24e\textsuperscript{-4}$}  & \textcolor{black}{Yes} \\
         Qatar           & $1.07e\textsuperscript{-5}$    & Yes   & \textcolor{black}{0.0401}  & No \\
         \textcolor{black}{Russia}          & \textcolor{black}{$3.21e\textsuperscript{-102}$}    & \textcolor{black}{Yes}   & \textcolor{black}{$1.40e\textsuperscript{-9}$}  & \textcolor{black}{Yes} \\
         Saudi Arabia    & $2.79e\textsuperscript{-6}$    & Yes   & \textcolor{black}{0.0091}  & No \\
         Uganda          & 0.0002                  & Yes   & \textcolor{black}{0.0392}  & No \\
        \bottomrule
    \end{tabular}
    \caption{Significance (Sig.) results for pairwise comparisons between target countries and United Kingdom.}
    \label{tab:significance_results}
\end{table}

\paragraph{Debias tuning further mitigates bias in multiple languages} Debias using personas' official languages showed similar, but improved trends, as seen in Figure \ref{fig:llama-fnr}, bottom row. For the \texttt{Llama 3.1} model, Afghanistan (66.97\%), Brunei (78.95\%), and Saudi Arabia (80.96\%) had high FNRs, while the United Kingdom (54.90\%) and the United States (55.10\%) fared better. \texttt{debias-llama} reduced FNRs significantly, with Afghanistan at 33.62\%, Saudi Arabia at 28.11\%, and the United States at 18.90\%, almost matching the baseline (18.72\%). \textcolor{black}{The two countries that showed limited improvement were Cuba and Nicaragua, where FNRs decreased, but to a smaller extent (from 79.77\% and 76.4\%, to 62.70\% and 62.76\%, respectively).} The \texttt{debias-llama-lang} model further improved performance in regions like Afghanistan (30.97\%) and Saudi Arabia (24.01\%), though Australia and Brunei remained high at 49.04\% and 44.65\%, respectively. \textcolor{black}{We observe the same pattern with Cuba and Nicaragua as with \texttt{debias-llama}, highlighting the model's difficulty in handling Spanish text when it has not been debiased for this language.} Still, these results highlight again debias tuning’s effectiveness in reducing geographic bias and improving fairness, even in multilingual contexts, \textcolor{black}{where the models could generalize to unseen languages, although results for some languages, like Spanish, could still be improved.}

\paragraph{Effectiveness of debias tuning on unincluded countries} In Figure \ref{fig:llama-fnr}, we see that the FNRs are reduced for countries not included in the debias tuning. For some countries, \textcolor{black}{such as Nigeria, North Korea, and Uganda}, the \texttt{Llama 3.1} model resulted in an FNR of around 62\%, while the \texttt{debias-llama} model lowered this to approx. 25\%. \textcolor{black}{But, for other countries, such as Cuba or Russia, the reduction was not as notable.} This highlights the effectiveness of our method, as disparities among countries are less pronounced despite these countries not being specifically included in the debiasing process. \textcolor{black}{Still, challenges remain to achieve a model were there are no big differences among countries.} This behaviour is consistent across both the debias models and the debias language models. 

\section{Error Analysis}
\label{sec:error_analysis}

While our method effectively reduces biases, the F1-scores remain at \textcolor{black}{around \num{0.79}} for \texttt{Llama 3.1} and \texttt{Nemo} when prompted with country context. \textcolor{black}{For Phi 4, a lighter, less powerfull model, the F1-scores remain at \num{0.60}.} To identify remaining classification issues, we conducted an error analysis on a sample of \num{100} instances per error type for \textcolor{black}{\texttt{Llama 3.1} and \texttt{Nemo}} models, \textcolor{black}{as these models showed the most promising performance for potential deployment}. This analysis focused on common misclassifications, examining both false negatives and false positives for the debias models on English texts. We used error categories from \citet{van-aken-etal-2018-challenges} to guide our evaluation.

\subsection{Error Classes of False Negatives}

\paragraph{Hate speech without swear words} Identifying implicit hate and hate speech without swear words is a well-known challenge \cite{davidsonhatespeech,piot2024decodinghateexploringlanguage}. In our manual evaluation, 14\% of the posts fall into this category, with the majority targeting women. This highlights the limitations of existing models, which often rely on overtly offensive language or specific keywords to detect hate speech. Developing models capable of accurately identifying implicit hate speech requires incorporating a deeper understanding of context, intent, and subtle biases present in language. Such improvements are crucial to effectively addressing these hidden forms of harm. 

\scriptsize
\begin{leftbubbles}
\textit{The women skaters can't fall and make it look graceful like the men}
\end{leftbubbles}
\normalsize

\paragraph{Rhetorical questions} A common strategy in hate speech is the use of rhetorical questions \cite{Parvaresh2023}, often indicated by multiple question marks or exclamation points. Although only 2\% of our manual sample exhibited this pattern, it poses a challenge for state-of-the-art models to detect and requires further investigation.

\scriptsize
\begin{leftbubbles}
\textit{Are there any good female comedians?? Or comediennes if you prefer?}
\end{leftbubbles}
\normalsize

\paragraph{Metaphors and comparisons} Another type of implicit hate speech involves the use of metaphors and comparisons. \citet{10.3233/SW-180338} noted that understanding these texts requires complex reasoning as well as cultural and social knowledge. In our subsample, 2\% of the posts followed this pattern. While this is not the most common type of error, we emphasise the importance of developing models that can effectively identify implicit hate speech in all its forms.

\scriptsize
\begin{leftbubbles}
\textit{Releasing private Sony e mails to hurt people is the same as releasing\\ nude photos of Jennifer Lawrence.  Why are they ok t...}
\end{leftbubbles}
\normalsize


\paragraph{Sarcasm and irony} Detecting sarcasm and irony is a well-known challenge in NLP, especially in hate speech, where they can mask hateful intent. In our manual sample, sarcasm and irony appear in 3\% of the posts. While not common, this poses a significant challenge as these strategies often convey the opposite of their literal meaning.

\scriptsize
\begin{leftbubbles}
\textit{I guess I would be a safer driver too if I did ten under the speed limit}
\end{leftbubbles} 
\normalsize

\paragraph{Doubtful labels} In this class, we include data points where we question whether the original label was correct, based on our definition of hate speech. For example, we noticed that phrases like ``I hate'' are often labelled as hate speech in the original dataset, even when the target did not fit the definition. In fact, 74\% of the posts we analysed fall into this category. This raises concerns about potentially incorrect labels, underscoring the challenges of annotating hate speech.

\scriptsize
\begin{leftbubbles}
\textit{I hate racist people very much.}
\end{leftbubbles}
\normalsize

\subsection{Error Classes of False Positives}

\paragraph{Regular use of swear words} When analysing false positives, we found that half of our sample falls into this category, which includes various non-hateful uses of language. This encompasses affectionate swear words, such as saying ``\textit{You are a badass!}'', as well as words like \textit{hate}, \textit{abuse}, or \textit{harassed} when used in non-hate contexts, as well as negative words such as \textit{dead} or \textit{gun} that appear in neutral or unrelated contexts. Additionally, instances of negative or critical statements, such as ``\textit{This is his worst performance}'', but misclassified as hate, are included. These examples highlight the importance of understanding context to differentiate hate from non-hate language.

\scriptsize
\begin{leftbubbles}
\textit{WTF score did you expect serving liver??}
\end{leftbubbles}
\normalsize

\paragraph{Quotations or references} A major challenge for hate speech detection models is distinguishing between actual hate speech and quotes or references to hate speech. These often include quotation marks or are explicitly contextualised, such as ``\textit{he said that...}''. In our analysis, we found that 14\% of the posts fall into this category. This highlights the need for models to better understand context and intent when identifying hate speech.

\scriptsize
\begin{leftbubbles}
\textit{Yes, and ``girls suck at basketball'' is such an unconventional sentiment.}
\end{leftbubbles} 
\normalsize

\paragraph{Idiosyncratic or rare words} In our manual evaluation, we found that 11\% of the posts fell into this category. These posts included rare words, misspellings, abbreviations, slang, words in other languages, or highly descriptive hashtags. This issue emphasises the need for hate speech detection models to improve their handling of less common language elements, as failing to do so can result in overlooking harmful content or mislabelling non-hateful expressions.

\scriptsize
\begin{leftbubbles}
\textit{What does small town India think about \#MeToo? Kaam ke liye sweetoo,\\ kaam ke baad \#MeToo !}
\end{leftbubbles} 
\normalsize

\paragraph{Doubtful labels} We found fewer doubtful labels compared to false negatives. In our sample, 8\% of the posts were categorised as doubtful labels, with most of them falling under the category of implicit hate. This suggests that implicit hate remains a challenge for accurate classification, even among false positives.

\scriptsize
\begin{leftbubbles}
\textit{call me sexist but I hate audiobooks read by women}
\end{leftbubbles} 
\normalsize

In summary, most false negatives fall under doubtful labels, underscoring the challenges of annotating hate speech. Conversely, false positives often stem from misclassifying regular swear words, highlighting the need for methods to differentiate between insults and hate speech. Additionally, 5\% of false negatives and 17\% of false positives are misclassifications that do not fit into established categories.

\section{Conclusions}

This study shows that LLMs' memory features can introduce personalisation based on demographic attributes, which affects sensitive topics like hate speech. By using location-specific personas to simulate this feature, we observed systematic differences in classification outcomes depending on geographical context. We also examined whether the models displayed bias based on the language used in prompts, finding that this also influenced their behaviour. To address these variations, we applied a fine-tuning strategy that incorporates a consistency-based penalty in the custom loss function. This method aligns predictions with and without country context, leading to improved F1 scores and reducing bias in country-specific settings, thereby enhancing model performance across different contexts. Additionally, we conducted a detailed error analysis of the misclassifications made by the debias models. Future research may investigate further debiasing techniques and expand this approach to other types of social bias that could arise from the memory personalisation features of LLMs, ultimately contributing to more inclusive and robust AI systems.

\section{Limitations}
Our study is limited to a small set of countries, which may not fully capture global cultural and linguistic diversity, reducing the generalisability of our findings. Additionally, we used \textcolor{black}{five} LLM architectures, limiting insight into how geographic bias varies across other models. \textcolor{black}{While our proposed approach improves consistency, it may reduce sensitivity to context-specific signals. We limited training to one epoch and monitored the loss to reduce overfitting.} The debiasing approach was tested solely on hate speech detection, leaving its efficacy for other tasks unexamined. Potential societal impacts include reinforcing bias if debiasing is ineffective or misused. To mitigate misuse, we advocate for transparency in model training and deployment and stress the importance of ongoing evaluation across diverse contexts.

\section*{Computational Resources}
Experiments were conducted using a private infrastructure, which has a carbon efficiency of 0.432 kgCO$_2$eq/kWh. A cumulative of 1600 hours of computation was performed on hardware of type RTX A6000. Total emissions are estimated to be 207.36 kgCO$_2$eq of which 0 percent were directly offset. Estimations were conducted using the MachineLearning Impact calculator presented in \citet{lacoste2019quantifying}. \textcolor{black}{Running LLMs, especially reasoning models, can be costly. While techniques like knowledge distillation help improve accessibility \cite{piot2024efficientexplainablehatespeech}, our goal is not to use LLMs as hate speech detectors. Rather, we study how they respond to or generate hate speech in conversational settings. By doing so, we evaluate their behaviour in realistic interactions, rather than treating them as classifiers.}



\bibliography{aaai25}

\end{document}